\documentclass[runningheads]{llncs}
\usepackage{graphicx}
\usepackage{amsmath,amssymb} 
\usepackage{color,subfigure,wrapfig}
\usepackage{float}
\usepackage{amsmath,amssymb} 
\usepackage{cleveref}
\usepackage{color}
\usepackage{cite,caption}

\begin{document}

\newcommand{\point}{
    \raise0.7ex\hbox{.}
    }


\pagestyle{headings}

\mainmatter

\title{phi-LSTM: A Phrase-based Hierarchical \\ LSTM Model for Image Captioning}  

\titlerunning{phi-LSTM: A Phrase-based Hierarchical \\ LSTM Model for Image Captioning} 

\authorrunning{Tan and Chan} 

\author{Ying Hua Tan and Chee Seng Chan}
\institute{Center of Image and Signal Processing, \\ Faculty of Computer Science \& Information Technology, \\ University of Malaya, Kuala Lumpur, Malaysia \\
\email{tanyinghua@siswa.um.edu.my; cs.chan@um.edu.my}
}

\maketitle

\begin{abstract}
{\it A picture is worth a thousand words}. Not until recently, however, we noticed some success stories in understanding of visual scenes: a model that is able to detect/name objects, describe their attributes, and recognize their relationships/interactions. In this paper, we propose a phrase-based hierarchical Long Short-Term Memory (phi-LSTM) model to generate image description. The proposed model encodes sentence as a sequence of combination of phrases and words, instead of a sequence of words alone as in those conventional solutions. The two levels of this model are dedicated to i) learn to generate image relevant noun phrases, and ii) produce appropriate image description from the phrases and other words in the corpus. Adopting a convolutional neural network to learn image features and the LSTM to learn the word sequence in a sentence, the proposed model has shown better or competitive results in comparison to the state-of-the-art models on Flickr8k and Flickr30k datasets. 
\end{abstract}

\section{Introduction}

\begin{wrapfigure}{r}{0.5\linewidth}
\centering
\captionsetup{justification=centering}
\vspace{-20pt}
\includegraphics[height=0.56\linewidth, width=0.8\linewidth]{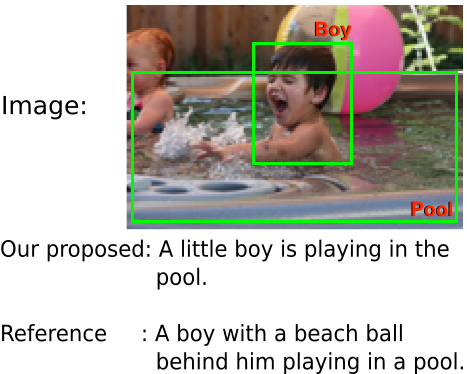}
\caption{Complete visual scene understanding is a holy grail in computer vision.}
\vspace{-20pt}
	\label{fig:intro}
\end{wrapfigure} Automatic caption/description generation from images is a challenging problem that requires a combination of visual information and linguistic as illustrated in Fig. \ref{fig:intro}. In other words, it requires not only complete image understanding, but also sophisticated natural language generation \cite{sadeghi2011recognition,gupta2012choosing,bernardi2016automatic,rasiwasia2010new}. This is what makes it such an interesting task that has been embraced by both the computer vision and natural language processing communities. 

One of the most common models applied for automatic caption generation is a neural network model that composes of two sub-networks \cite{mao2014deep,Vinyals2015,karpathy2015deep,kiros2014unifying,donahue2015long, xu2015show}, where a convolutional neural network (CNN) \cite{krizhevsky2012imagenet} is used to obtain feature representation of an image; while a recurrent neural network (RNN)\footnote{RNN is a popular choice due to its capability to process arbitrary length sequences like language where words sequence governing its semantic is order-sensitive.} is applied to encode and generate its caption description. In particular, Long Short-Term Memory (LSTM) model \cite{hochreiter1997long} has emerged as the most popular architecture among RNN, as it has the ability to capture long-term dependency and preserve sequence. Although sequential model is appropriate for processing sentential data, it does not capture any other syntactic structure of language at all. Nevertheless, it is undeniable that sentence structure is one of the prominent characteristics of language, and Victor Yngve - an influential contributor in linguistic theory stated in 1960 that``\textit{language structure involving, in some form or other, a phrase-structure hierarchy, or immediate constituent organization}''\cite{yngve1960model}. Moreover, Tai et al. \cite{tai2015improved} proved that a tree-structured LSTM model that incorporates syntactic interpretation of sentence structure, can learn the semantic relatedness between sentences better than a pure sequential LSTM alone. This gives rise to question of whether is it a good idea to disregard other syntax of language in the task of generating image description.

\begin{figure}[t]
		\centering
		\subfigure[]{
		\includegraphics[height=0.4\linewidth, width=0.45\linewidth]{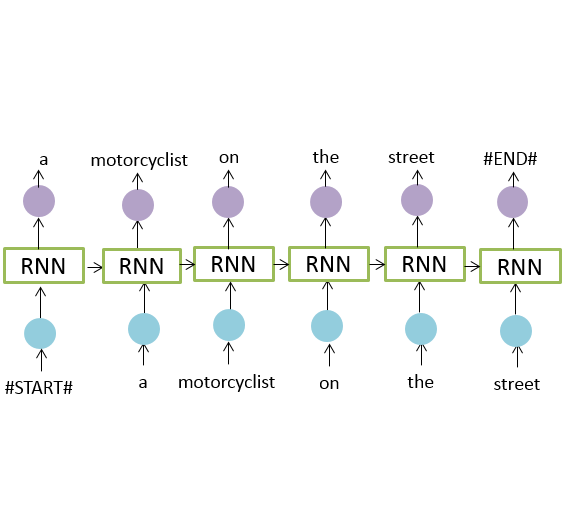}
		\label{fig:RNN}
	}
	\subfigure[]{
		\includegraphics[height=0.4\linewidth, width=0.5\linewidth]{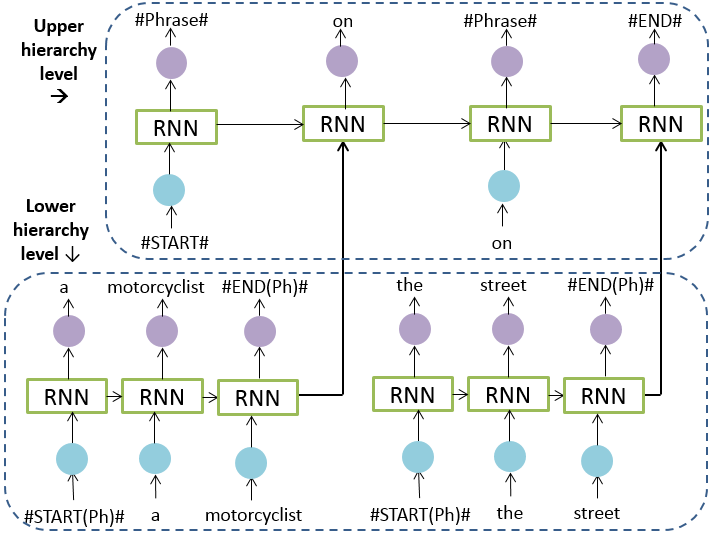}
		\label{fig:ph-lstm}
	}
	\subfigure{
	\includegraphics[width=0.5\linewidth]{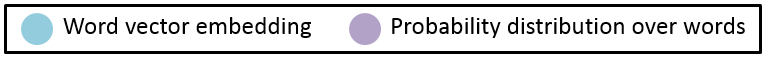}
	}
	\caption{Model comparison: (a) Conventional RNN language model, and (b) our proposed phrase-based model.}
	\label{fig:comparison}
\end{figure}

In this paper, we would like to investigate the capability of a phrase-based language model in generating image caption as compared to the sequential language model such as \cite{Vinyals2015}. To this end, we design a novel phrase-based hierarchical LSTM model, namely {\bf phi-LSTM} to encode image description in three stages - chunking of training caption, image-relevant phrases composition as a vector representation and finally, sentence encoding with image, words and phrases. As opposed to those conventional RNN language models which process sentence as a sequence of words, our proposed method takes noun phrase as a unit in the sentence, and thus processes the sentential data as a sequence of combination of both words and phrases together. Fig. \ref{fig:comparison} illustrates the difference between the conventional RNN language model and our proposal with an example. Both phrases and sentences in our proposed model are learned with two different sets of LSTM parameters, each models the probability distribution of word conditions on previous context and image. Such design is motivated by the observation that some words are more prone to appear in phrase, while other words are more likely to be used to link phrases. In order to train the proposed model, a new perplexity based cost function is defined. Experimental results using two publicly available datasets (Flickr8k \cite{rashtchian2010collecting} and Flickr30k \cite{young2014image}), and a comparison to the state-of-the-art results \cite{karpathy2015deep,Vinyals2015,mao2014deep,donahue2015long,lebret2015phrase} have shown the efficacy of our proposed method.

\section{Related Works}
The image description generation task is generally inspired by two lines of research, which are (i) the learning of cross-modality transition or representation between image and language, and (ii) the description generation approaches.

\subsection{Multimodal Representation and Transition} 
To model the relationship between image and language, some works associate both modalities by embedding their representations into a common space \cite{hodosh2013framing, frome2013devise, socher2014grounded, karpathy2014deep}. First, they obtain the image features using a visual model like CNN \cite{frome2013devise, socher2014grounded}, as well as the representation of sentence with a language model such as recursive neural network \cite{socher2014grounded}. Then, both of them are embedded into a common multimodal space and the whole model is learned with ranking objective for image and sentence retrieval task. This framework was also tested at object level by Karpathy et al. \cite{karpathy2014deep} and proved to yield better results for the image and sentence bi-directional retrieval task. Besides that, there are works that learn the probability density over multimodal inputs using various statistical approaches. These include Deep Boltzmann Machines \cite{srivastava2012multimodal}, topic models \cite{jia2011learning}, log-bilinear neural language model \cite{kiros2014multimodal, kiros2014unifying} and recurrent neural networks \cite{karpathy2015deep, mao2014deep, Vinyals2015} etc. Such approaches fuse different input modalities together to obtain a unified representation of the inputs. It is notable to mention that there are also some works which do not explicitly learn the multimodal representation between image and language, but transit between modalities with retrieval approach. For example, Kuznetsova et al. \cite{kuznetsova2014treetalk} retrieve images similar to the query image from their database, and extract useful language segments (such as phrases) from the descriptions of the retrieved images.

\subsection{Description Generation}
On the other hand, caption generation approaches can generally be grouped into three categories as below:
\subsubsection{Template-based.}
These approaches generate sentence from a fixed template \cite{farhadi2010every, kulkarni2011baby, yang2011corpus, mitchell2012midge, gupta2012image}. For example, Farhadi et al. \cite{farhadi2010every} infer a single triplet of object, action and scene from an image and convert it into a sentence with fixed template. Kulkarni et al. \cite{kulkarni2011baby} use complex graph of detections to infer elements in sentence with conditional random field (CRF), but the generation of sentences is still based on the template. Mitchell et al. \cite{mitchell2012midge} and Gupta et al. \cite{gupta2012image} use a more powerful language parsing model to produce image description. In overall, all these approaches generate description which is syntactically correct, but rigid and not flexible.
\subsubsection{Composition Method.}
These approaches extract components related to the images and stitch them up to form a sentence \cite{li2011composing, kuznetsova2012collective, kuznetsova2014treetalk}. Description generated in such manner is broader and more expressive compared to the template-based approach, but is more computationally expensive at test time due to its non-parametric nature.
\subsubsection{Neural Network.}
These approaches produce description by modeling the conditional probability of a word given multimodal inputs. For instance, Kiros et al. \cite{kiros2014multimodal,kiros2014unifying} developed multimodal log-bilinear neural language model for sentence generation based on context and image feature. However, it has a fixed window context. The other popular model is recurrent neural network \cite{karpathy2015deep, mao2014deep, chen2014learning, Vinyals2015,donahue2015long}, due to its ability to process arbitrary length of sequential inputs such as sequence of words. This model is usually connected with a deep CNN that generates image features. The variants on how this sub-network is connected to the RNN have been investigated by different researchers. For instance, the multimodal recurrent neural network proposed by Mao et al. \cite{mao2014deep} introduces a multimodal layer at each time step of the RNN, before the softmax prediction of words. Vinyals et al. \cite{Vinyals2015} treat the sentence generation task as a machine translation problem from image to English, and thus image feature is employed in the first step of the sequence trained with their LSTM RNN model.
\subsection{Relation to Our Work}
Automatic image caption generated via template-based \cite{farhadi2010every, kulkarni2011baby, yang2011corpus, mitchell2012midge, gupta2012image} and composition methods \cite{li2011composing, kuznetsova2012collective, kuznetsova2014treetalk} are typically two-stage approaches, where relevant elements such as objects (noun phrases) and relations (verb and prepositional phrases) are generated first before a full descriptive sentence is formed with the phrases. With the capability of LSTM model in processing long sequence of words, neural network based method that uses a two-stage approach deem unnecessary. However, we are still interested to find out how sequential model with phrase as a unit of sequence performs. The closest work related to ours is the one proposed by Lebret et al. \cite{lebret2015phrase}. They obtain phrase representation with simple word vector addition and learn its relevancy with image by training with negative samples. Sentence is then generated as a sequence of phrases, predicted using a statistical framework conditioned on previous phrases and its chunking tags. While their aim was to design a phrase-based model that is simpler than RNN, we intend to compare RNN phrase-based model with its sequential counterpart. Hence, our proposed model generates phrases and recomposes them into sentence with two sub-networks of LSTM, which are linked to form a hierarchical structure as shown in Fig. \ref{fig:ph-lstm}.

\section{Our Proposed phi-LSTM Model}
\label{sec: model}
This section details how the proposed method encodes image description in three stages - i) chunking of image description, ii) encode words and phrases into distributed representations, and finally iii) encodes sentence with the phi-LSTM model.

\subsection{Phrase Chunking}

\begin{figure}[t]
	\centering
\includegraphics[height=0.35\linewidth, width=0.95\linewidth]{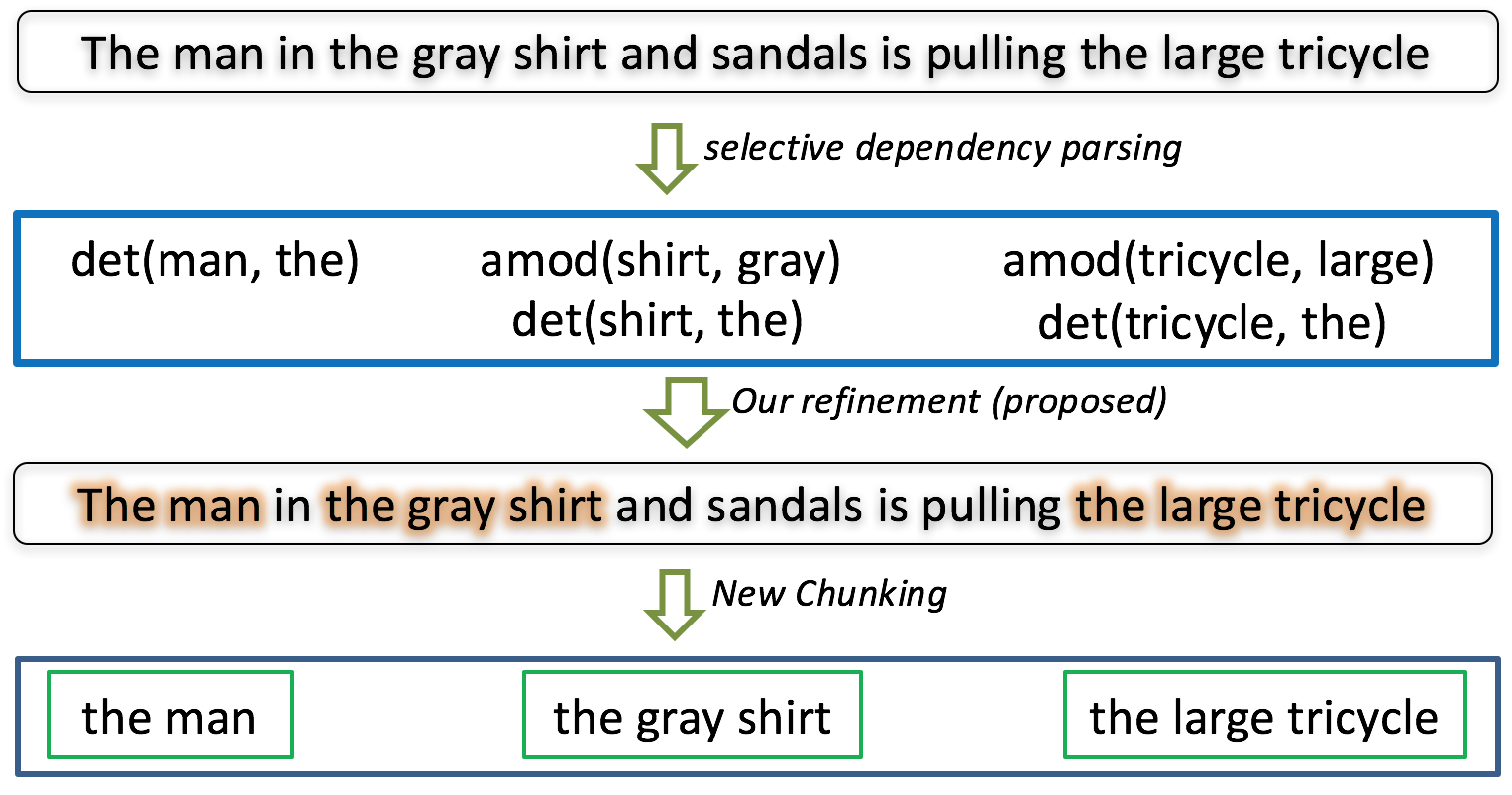}
\caption{Phrase chunking from dependency parse.}
	\label{fig:phrase}
\end{figure}

A quick overview on the structure of image descriptions reveals that, key elements which made up the majority of captions are usually noun phrases that describe the content of the image, which can be either objects or scene. These elements are linked with verb and prepositional phrases. Thus, noun phrase essentially covers over half of the corpus in a language model trained to generate image description. And so, in this paper, our idea is to partition the learning of noun phrase and sentence structure so that they can be processed more evenly, compared to extracting all phrases without considering their part of speech tag. 

To identify noun phrases from a training sentence, we adopt the dependency parse with refinement using Stanford CoreNLP tool \cite{manning-EtAl:2014:P14-5}, which provides good semantic representation over a sentence by providing structural relationships between words. Though it does not chunk sentence directly as in constituency parse and other chunking tools, the pattern of noun phrase extracted is more flexible as we can select desirable structural relations. The relations we selected are:

\begin{itemize}
	\item determiner relation (\textit{det}), 
	\item numeric modifier (\textit{nummod}), 
	\item adjectival modifier (\textit{amod}), 
	\item adverbial modifier (\textit{advmod}), but is selected only when the meaning of adjective term is modified, e.g. ``\textit{dimly lit room}", 
	\item compound (\textit{compound}), 
	\item nominal modifier for possessive alteration (\textit{nmod:of} \& \textit{nmod:poss}). 
\end{itemize}

Note that the dependency parse only extracts triplet made up of a governor word and a dependent word linked with a relation. So, in order to form phrase chunk with the dependency parse, we made some refinements as illustrated in Fig.  \ref{fig:phrase}. The triplets of selected relations in a sentence are first located, and those consecutive words (as highlighted in the figure, e.g. ``the", ``man'') are grouped as a single phrase, while the standalone word (e.g. ``in") will remain as a unit in the sentence.

\subsection{Compositional Vector Representation of Phrase}
This section describes how compositional vector representation of a phrase is computed, given an image. 

\subsubsection{Image Representation.}
A 16-layer VggNet \cite{simonyan2014very} pre-trained on ImageNet \cite{deng2009imagenet} classification task is applied to learn image feature in this work. Let $\mathbf{I} \in \mathbb{R}^D$  be an image feature, 
it is embedded into a K-dimensional vector, $\mathbf{v_p}$ with image embedding matrix, $\mathbf{W_{ip}} \in \mathbb{R}^{K \times D}$ and bias $\mathbf{b_{ip}} \in \mathbb{R}^K$. 
\begin{equation}
\mathbf{v_p} = \mathbf{W_{ip}} \mathbf{I} + \mathbf{b_{ip}}~.
\end{equation}

\subsubsection{Word Embedding.}
Given a dictionary $\mathcal{W}$ with a total of  \textit{V} vocabulary, where word $\mathit{w} \in \mathcal{W}$ denotes word in the dictionary, a word embedding matrix $\mathbf{W_e} \in \mathbb{R}^{K \times V}$ is defined to encode each word into a \textit{K}-dimensional vector representation, \textbf{x}. Hence, an image description with words $\mathit{w}_1 \cdots \mathit{w}_M$ will correspond to vectors $\mathbf{x}_1 \cdots \mathbf{x}_M$ accordingly.

\begin{figure}[t]
	\centering
	\includegraphics[height=0.3\linewidth, width=0.7\linewidth]{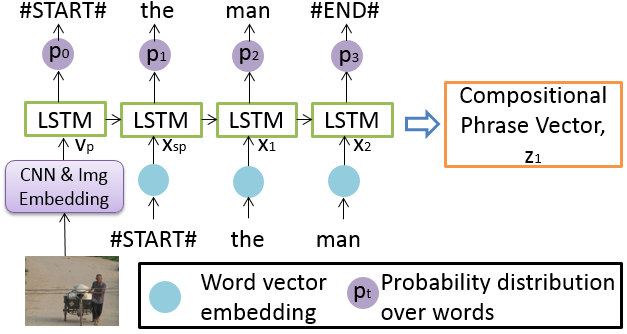}
	\caption{Composition of phrase vector representation in phi-LSTM.}
	\label{fig:phrase_gen}
\end{figure}

\subsubsection{Composition of Phrase Vector Representation.}
For each phrase extracted from the sentence, a LSTM-based RNN model similar to \cite{Vinyals2015} is used to encode its sequence as shown in Fig. \ref{fig:phrase_gen}. Similar to \cite{Vinyals2015}, we treat the sequential modeling from image to phrasal description as a machine translation task, where the embedded image vector is inputted to the RNN on the first time step, followed by a start token $\mathbf{x_{sp}} \in \mathbb{R}^{K}$ indicating the translation process. It is trained to predict the next word at each time step by outputting  $\mathbf{p_{{t_p}+1}} \in \mathbb{R}^{K \times V}$, which is modeled as the probability distribution over all words in the corpus. The last word of the phrase will predict an end token. So, given a phrase \textit{P} which is made up by \textit{L} words, the input $\mathbf{x_{t_p}}$ at each time step are:
\begin{equation}
\mathbf{x_{t_p}} =
	\begin{cases}
	\mathbf{v_p}~, & \text{if}\ t_p=-1 \\
	\mathbf{x_{sp}}~, & \text{if}\ t_p=0 \\
	\mathbf{W_e}w_{t_{p}}~, & \text{for}\ t_p = {1...L}~.
	\end{cases}
\end{equation}

For a LSTM unit at time step $t_p$, let $\mathbf{i}_{t_p}, \mathbf{f}_{t_p}, \mathbf{o}_{t_p}, \mathbf{c}_{t_p}$ and $\mathbf{h}_{t_p}$ denote the input gate, forget gate, output gate, memory cell and hidden state at the time step respectively. Thus, the LSTM transition equations are:
\begin{equation}
\label{eq_lstm1}
\mathbf{i}_{t_p} = \sigma(\mathbf{W_i} \mathbf{x}_{t_p} + \mathbf{U_i} \mathbf{h}_{{t_p}-1})~,
\end{equation}
\begin{equation}
\label{eq_lstm2}
\mathbf{f}_{t_p} = \sigma(\mathbf{W_f} \mathbf{x}_{t_p} + \mathbf{U_f} \mathbf{h}_{{t_p}-1})~,
\end{equation}
\begin{equation}
\label{eq_lstm3}
\mathbf{o}_{t_p} = \sigma(\mathbf{W_o} \mathbf{x}_{t_p} + \mathbf{U_o} \mathbf{h}_{{t_p}-1})~,
\end{equation}
\begin{equation}
\label{eq_lstm4}
\mathbf{u}_{t_p} = tanh(\mathbf{W_u} \mathbf{x}_{t_p} + \mathbf{U_u} \mathbf{h}_{{t_p}-1})~,
\end{equation}
\begin{equation}
\label{eq_lstm5}
\mathbf{c}_{t_p} = \mathbf{i}_{t_p} \odot \mathbf{u}_{t_p} + \mathbf{f}_{t_p} \odot \mathbf{c}_{{t_p}-1}~,
\end{equation}
\begin{equation}
\label{eq_lstm6}
\mathbf{h}_{t_p} = \mathbf{o}_{t_p} \odot tanh(\mathbf{c}_{t_p})~,
\end{equation}
\begin{equation}
\label{eq_lstm7}
\mathbf{p}_{{t_p}+1} = \text{softmax}(\mathbf{h}_{t_p})~.
\end{equation}

Here, $\sigma$ denotes a logistic sigmoid function while $\odot$ denotes elementwise multiplication. The LSTM parameters \{$\mathbf{W_i}, \mathbf{W_f}, \mathbf{W_o}, \mathbf{W_u}, \mathbf{U_i}, \mathbf{U_f}, \mathbf{U_o}, \mathbf{U_u}$\} are all matrices with dimension of $\mathbb{R}^{K \times K}$. Intuitively, each gating unit controls the extent of information updated, forgotten and forward-propagated while the memory cell holds the unit internal memory regarding the information processed up to current time step. The hidden state is therefore a gated, partial view of the memory cell of the unit. At each time step, the probability distribution of words outputted is equivalent to the conditional probability of word given the previous words and image, $P(w_t | w_{1:t-1},I)$. On the other hand, the hidden state at the last time step \textit{L} is used as the compositional vector representation of the phrase, $\mathbf{z} \in \mathbb{R}^K$ , where $\mathbf{z} = \mathbf{h}_{L}$.

\subsection{Encoding of Image Description}

\begin{figure}[t]
	\centering
	\includegraphics[height=0.3\linewidth, width=0.9\linewidth]{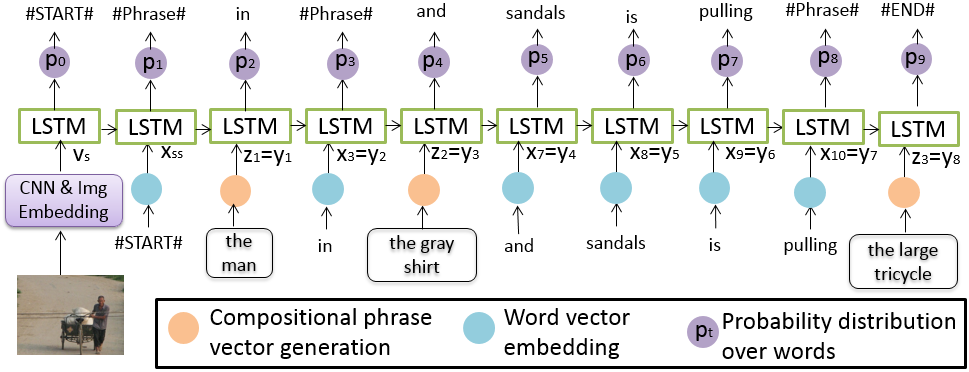}
	\caption{Sentence encoding with phi-LSTM.}
	\label{fig:sent_RNN}
\end{figure}

Once the compositional vector of phrases are obtained, they are linked with the remaining words in the sentence using another LSTM-based RNN model as shown in Fig. \ref{fig:sent_RNN}. Another start token $\mathbf{x_{ss}} \in \mathbb{R}^{K}$ and image representation $\mathbf{v_s} \in \mathbb{R}^{K}$ are introduced, where
\begin{equation}
\mathbf{v_s} = \mathbf{W_{is}} \mathbf{I} + \mathbf{b_{is}}~,
\end{equation}
with $\mathbf{W_{is}} \in \mathbb{R}^{K \times D}$ and bias $\mathbf{b_{is}} \in \mathbb{R}^K$ as embedding parameters. Hence, the input units of the LSTM in this level will be the image representation $\mathbf{v_s}$, start token $\mathbf{x_{ss}}$, followed by either compositional vector of phrase \textbf{z} or word vector \textbf{x} in accordance to the sequence of its description.

For simplicity purpose, the arranged input sequence will be referred as \textbf{y}. Therefore, given the example in Fig. \ref{fig:phrase_gen}-\ref{fig:sent_RNN}, the LSTM input sequence of the sentence will be \{$\mathbf{v_s}, \mathbf{x_{ss}}, \mathbf{y}_1 ... \mathbf{y}_N$\} where \textit{N} = 8, and it  is equivalent to sequence \{$\mathbf{v_s}, \mathbf{x_{ss}}, \mathbf{z}_1, \mathbf{x}_3, \mathbf{z}_2, \mathbf{x}_7, \mathbf{x}_8, \mathbf{x}_9, \mathbf{x}_{10}, \mathbf{z}_3$\}, as in Fig. \ref{fig:sent_RNN}. Note that a phrase token is added to the vocabulary, so that the model can predict it as an output when the next input is a noun phrase.

The encoding of the sentence is similar to the phrase vector composition. Eq. \ref{eq_lstm1}-\ref{eq_lstm7} are applied here using $\mathbf{y_{t_s}}$ as input instead of $\mathbf{x_{t_p}}$, where $t_p$ and $t_s$ represent time step in phrase and sentence respectively. A new set of model parameters with same dimensional size is used in this hierarchical level. 

\section{Training the phi-LSTM Model}
\label{sec: training}
The proposed phi-LSTM model is trained with log-likelihood objective function computed from the perplexity\footnote{ Perplexity is a standard approach to evaluate language model.} of sentence conditioned on its corresponding image in the training set. Given an image \textbf{I} and its description \textbf{S}, let \textit{R} be the number of phrases of the sentence, $ P_i $ correspond to the number of LSTM blocks processed to get the compositional vector of phrase \textit{i}, \textit{Q} is the length of composite sequence of sentence \textbf{S}, while $\mathbf{p_{t}}_{p}$ and $\mathbf{p_{t}}_{s}$ are the probability output of LSTM block at time step $\mathit{t_{p}}-1$ and $\mathit{t_{s}}-1$ for phrase and sentence level respectively. The perplexity of sentence \textbf{S} given its image \textbf{I} is
\begin{equation}
\log_2 \mathcal{PPL}(\mathbf{S}|\mathbf{I}) = -\frac{1}{N} \left[ \sum_{t_{s}=-1}^{Q} \log_2 \mathbf{p_t}_{s} + \sum_{i=1}^{R} \left[\sum_{t_{p}=-1}^{P_i} \log_2 \mathbf{p_t}_{p} \right] \right]~,
\label{eq: cost1}
\end{equation}
where
\begin{equation}
N = Q + \sum_{i=1}^{R} P_i~.
\end{equation}
Hence, with \textit{M} number of training samples, the cost function of our model is: 
\begin{equation}
\mathcal{C(\theta)} = -\frac{1}{L} \sum_{j=1}^{M} \left[N_j \log_2 \mathcal{PPL} (\mathbf{S_j}|\mathbf{I_j}) \right] + \lambda_\theta \cdot \parallel \theta \parallel_2^2~,
\end{equation}
where
\begin{equation}
L = M\times\sum_{j=1}^{M} N_j~.
\end{equation}
It is the average log-likelihood of word given their previous context and the image described, summed with a regularization term, $\lambda_\theta \cdot \parallel \theta \parallel_2^2$, average over the number of training samples. Here, $\theta$ is the parameters of the model. 

This objective however, does not discern on the appropriateness of different inputs at each time step. So, given multiple possible inputs, it is unable to distinguish which phrase is the most probable input at that particular time step during the decoding stage. That is, when a phrase token is inferred as the next input, all possible phrases will be inputted in the next time step. The candidate sequences are then ranked according to their perplexity up to this time step, where only those with high probability are kept. Unfortunately, this is problematic because subject in an image usually has much lower perplexity as compared to object and scene. Thus, such algorithm will end up generating description made up of only variants of subject noun phrases.

\begin{figure}[t]
	\centering
	\includegraphics[height=0.35\linewidth, width=0.9\linewidth]{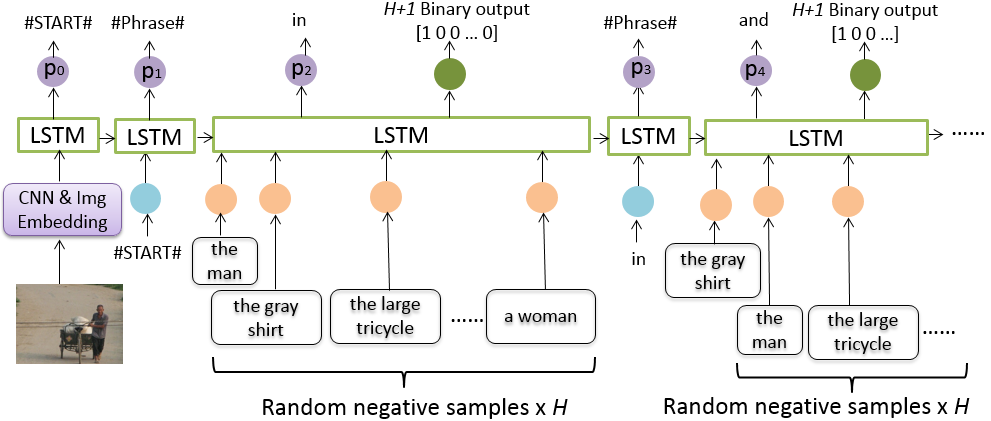}
	\caption{Upper hierarchy of phi-LSTM with phrase selection objective.}
	\label{fig:phrase_selection}
\end{figure}

To overcome this limitation, we introduce a phrase selection objective during the training stage. At all time steps when an input is a phrase, \textit{H} number of randomly selected phrases that are different from the ground truth input is feed into the phi-LSTM model as shown in Fig. \ref{fig:phrase_selection}. The model will then produce two outputs, which are the next word prediction solely based on the actual input, and a classifier output that distinguishes the actual one from the rest. Though the number of inputs at these time steps increases, the memory cell and hidden state that is carried to the next time step keep only information of the actual input. The cost function for phrase selection objective of a sentence is
\begin{equation}
\mathcal{C}_{PS} = \sum_{t_{s} \in \mathcal{P}} \sum_{k=1}^{H+1} \kappa_{t_{s}k} \sigma(1-y_{t_{s}k}h_{t_{s}k}\mathbf{W_{ps}})~.
\end{equation}
where $\mathcal{P}$ is the set of all time steps where the input is phrase, $h_{t_{s}k}$ is the hidden state output at time step $t_{s}$ from input \textit{k}, and $y_{t_{s}k}$ is its label which is +1 for the actual input and -1 for the false inputs. $\mathbf{W_{ps}} \in \mathbb{R}^{K \times 1}$ is trainable parameters for the classifier while $\kappa_{t_{s}k}$ scales and normalizes the objective based on the number of actual and false inputs at each time step. The overall objective function is then
\begin{equation}
\mathcal{C}_{F}(\theta) = -\frac{1}{L} \sum_{j=1}^{M} \left[N_j \log_2 \mathcal{PPL} (\mathbf{S_j}|\mathbf{I_j}) + \mathcal{C}_{PSj} \right] + \lambda_\theta \cdot \parallel \theta \parallel_2^2~.
\end{equation}

This cost function is minimized and backpropagated with RMSprop optimizer \cite{rmsprop} and trained in a minibatch of 100 image-sentence pair per iteration. We cross-validate the learning rate and weight decay depending on dataset, and dropout regularization \cite{srivastava2014dropout} is employed over the LSTM parameters during training to avoid overfitting. 

\section{Image Caption Generation}
\label{sec: captioning}
Generation of textual description using the phi-LSTM model given an image is similar to other statistical language models, except that the image relevant phrases are generated first in the lower hierarchical level of the proposed model. Here, embedded image feature of the given image followed by the start token of phrase are inputted into the model, acting as the initial context required for phrase generation. Then, the probability distribution of the next word over the vocabulary is obtained at each time step given the previous contexts, and the word with the maximum probability is picked and fed into the model again to predict the subsequent word. This process is repeated until the end token for phrase is inferred. As we usually need multiple phrases to generate a sentence, beam search scheme is applied and the top \textit{K} phrases generated are kept as the candidates to form the sentence. To generate a description from the phrases, the upper hierarchical level of the phi-LSTM model is applied in a similar fashion. When a phrase token is inferred, \textit{K} phrases generated earlier are used as the inputs for the next time step. Keeping only those phrases which generate positive result with the phrase selection objective, inference on the next word given the previous context and the selected phrases is performed again. This process iterates until the end token is inferred by the model.

Some constraints are added here, which are i) each predicted phrase may only appears once in a sentence, ii) maximum number of unit (word or phrase) that made up a sentence is limited to 20, iii) maximum number of words forming a phrase is limited to 10, and iv) generated phrases with perplexity higher than threshold \textit{T} are discarded.

\section{Experiment}
\label{sec: experiment}
\subsection{Datasets}
The proposed phi-LSTM model is tested on two benchmark datasets - Flickr8k \cite{rashtchian2010collecting} and Flickr30k \cite{young2014image}, and compared to the state-of-the-art methods \cite{karpathy2015deep,mao2014deep,Vinyals2015,donahue2015long,lebret2015phrase}. These datasets consist of 8000 and 31000 images respectively, each annotated with five ground truth descriptions from crowd sourcing. For both datasets, 1000 images are selected for validation and another 1000 images are selected for testing; while the rest are used for training. All sentences are converted to lower case, with frequently occurring punctuations removed and word that occurs less than 5 times (Flickr8k) or 8 times (Flickr30k) in the training data discarded. The punctuations are removed so that the image descriptions are consistent with the data shared by Karpathy and L. Fei-Fei \cite{karpathy2015deep}.

\subsection{Results Evaluated with Automatic Metric}
Sentence generated using the phi-LSTM model is evaluated with automatic metric known as the bilingual evaluation understudy (BLEU) \cite{papineni2002bleu}. It computes the n-gram co-occurrence statistic between the generated description and multiple reference sentences by measuring the n-gram precision quality. It is the most commonly used metric in this literature.

\begin{table}[t]
	\caption{BLEU score of generated sentence on (a) Flickr8k and (b) Flickr30k dataset.}
	\label{tab:results_bleu}
	\centering
\subtable[]{	
	\begin{tabular*}{0.43\linewidth}{l|*{4}{c}}
		\hline
		\multicolumn{5}{c}{Flickr8k} \\
		\hline
		Models & B-1 & B-2 & B-3 & B-4 \\
		\hline
		DeepVs \cite{karpathy2015deep} & 57.9 & 38.3 & 24.5 & 16.0 \\
		NIC	\cite{Vinyals2015} \protect\footnotemark & 60.2(63) & 40.4 & 25.9 & 16.5 \\
		phi-LSTM & \textbf{63.6} & \textbf{43.6} & \textbf{27.6} & \textbf{16.6} \\
		\hline
	\end{tabular*}
}
\subtable[]{	
	\begin{tabular*}{0.43\linewidth}{l|*{4}{c}}	
		\hline
		\multicolumn{5}{c}{Flickr30k} \\
		\hline
		Models & B-1 & B-2 & B-3 & B-4 \\
		\hline
		DeepVS \cite{karpathy2015deep}	& 57.3 & 36.9 & 24.0 & 15.7 \\
		mRNN \cite{mao2014deep}			& 60 & 41 & 28 & \textbf{19} \\
		NIC	\cite{Vinyals2015} \protect\footnotemark & 66.3(66) & 42.3 & 27.7 & 18.3 \\
		LRCNN \cite{donahue2015long}	& 58.7 & 39.1 & 25.1 & 16.5 \\
		PbIC \cite{lebret2015phrase}	& 59 & 35 & 20 & 12 \\
	phi-LSTM	& \textbf{66.6} & \textbf{45.8} & \textbf{28.2} & 17.0  \\
		\hline
	\end{tabular*}
}	
\end{table}

\addtocounter{footnote}{-1}
\footnotetext{The BLEU score reported here is computed on our implementation of NIC \cite{Vinyals2015}, and the bracketed value is the reported score by the author.} 
\addtocounter{footnote}{1}
\footnotetext{The BLEU score reported here is cited from \cite{karpathy2015deep}, and the bracketed value is the reported score by the author.} 

Table \ref{tab:results_bleu} shows the performance of our proposed model in comparison to the current state-of-the-art methods. NIC \cite{Vinyals2015} which is used as our baseline is a reimplementation, and thus its BLEU score reported here is slightly different from the original work. Our proposed model performs better or comparable to the state-of-the-art methods on both Flickr8k and Flickr30k datasets. In particular, we outperform our baseline on both datasets, as well as PbIC \cite{lebret2015phrase} - a work that is very similar to us on Flickr30k dataset by at least 5-10\%.

\begin{figure}[t]
	\centering
	\includegraphics[height=0.6\linewidth, width=0.9\linewidth]{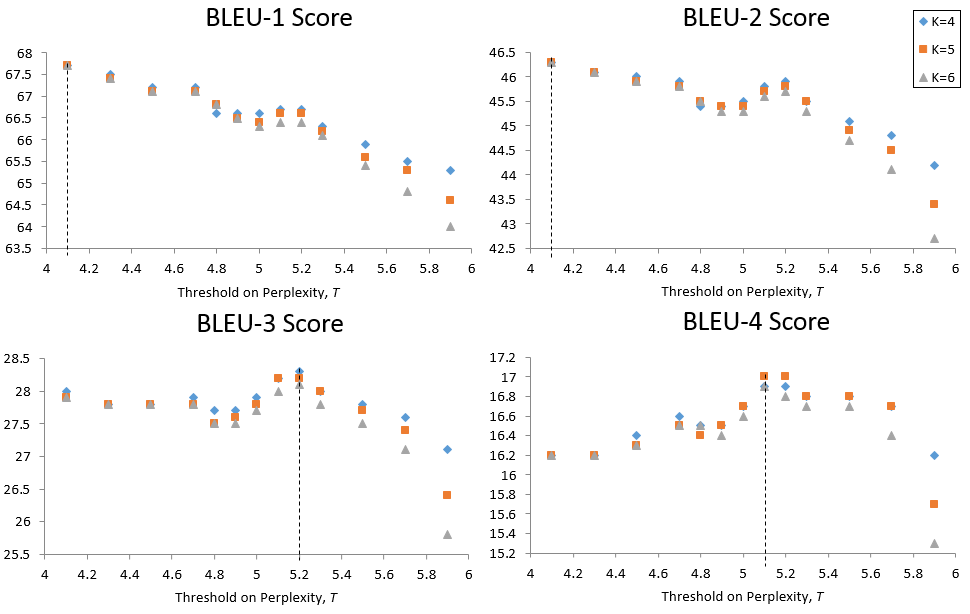}
	\caption{Effect of perplexity threshold \textit{T} and maximum number of phrases used for generating sentence, \textit{K} on BLEU score.}
	\label{fig:thres}
\end{figure}

\begin{figure}[t]
	\centering
	\includegraphics[height=0.25\linewidth, width=\linewidth]{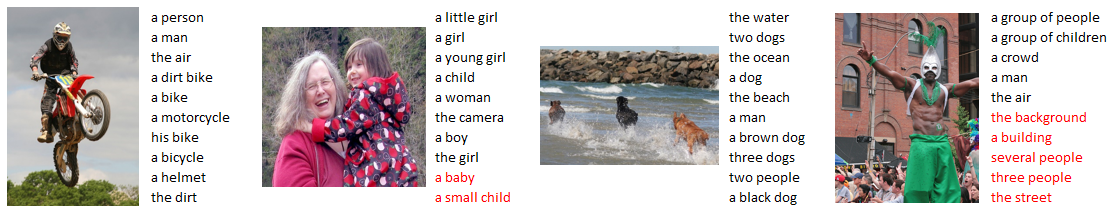}
	\caption{Example of phrases generated from image by lower hierarchical level of phi-LSTM model. Red fonts indicate that the log probability of that phrases is below threshold.}
	\label{fig:eg_phrases}
\end{figure}

As mentioned in Section \ref{sec: captioning}, we generate \textit{K} phrases from each image and discard those with perplexity higher than a threshold value \textit{T}, when generating the image caption. In order to understand how these two parameters affect our generated sentence, we use different \textit{K} and \textit{T} to generate the image caption with our proposed model trained on the Flickr30k dataset. Changes of the BLEU score against \textit{T} and \textit{K} are plotted in Fig. \ref{fig:thres}. It is shown that \textit{K} does not have a significant effect on the BLEU score, when \textit{T} is set to below 5.5. On the other hand, unigram and bi-gram BLEU scores improve with lower perplexity threshold, in contrast to tri-gram and 4-gram BLEU scores that reach an optimum value when \textit{T}=5.2. This is because the initial (few) generated phrases with the lowest perplexity are usually different variations of phrase describing the same entity, such as `\textit{a man}' and `\textit{a person}'. Sentence made with only such phrases has higher chance to match with the reference descriptions, but it would hardly get a match on tri-gram and 4-gram. In order to avoid generating caption made from only repetition of similar phrases, we select \textit{T} and \textit{K} which yield the highest 4-gram BLEU score, which are \textit{T}=6.5 and \textit{K}=6 on Flickr8k dataset, and \textit{T}=5.2 and \textit{K}=5 on Flickr30k dataset. A few examples are shown in Fig. \ref{fig:eg_phrases}.

\subsection{Comparison of phi-LSTM with Its Sequence Model Counterpart}

\begin{table}[t]
	\centering
	\caption{Vocab size, word occurrence and average caption length in train data, test data, and generated description on Flickr8k dataset.}
	\label{tab:word_statistic}	
	\begin{tabular*}{\linewidth}{l|c|c|c|c|c|c|c|c}
		\hline
		{} & \multicolumn{2}{c|}{\textbf{Train Data}} & \multicolumn{4}{c|}{\textbf{Test Data}} & \multicolumn{2}{c}{\textbf{Gen. Caption}} \\
		\hline
		Number of sentence & \multicolumn{2}{c|}{30000} & \multicolumn{2}{c|}{5000}  & \multicolumn{2}{c|}{1000} & \multicolumn{2}{c}{1000} \\
		\hline
		{} & Actual & Trained & Actual & Trained & Actual & Trained & NIC \cite{Vinyals2015} & phi-LSTM \\
		\hline
		Size of vocab & 7371 & 2538 & 3147 & 1919 & 1507 & 1187 & 128 & 154 \\
		Number of words & 324481 & 316423 & 54335 & 52683 & 11139 & 10806 & 8275 & 6750 \\
		Avg. caption length & 10.8 & 10.5 & 10.9 & 10.5 & 11.1 & 10.8 & 8.3 & 6.8 \\
		\hline
	\end{tabular*}
\end{table}

To compare the differences between a phrase-based hierarchical model and a pure sequence model in generating image caption, the phi-LSTM model and NIC \cite{Vinyals2015} are both implemented using the same training strategy and parameter tuning. We are interested to know how well the corpus is trained by both models. Using the Flickr8k dataset, we computed the corpus information of i) the training data, ii) the reference sentences in the test data and iii) the generated captions as tabulated in Table \ref{tab:word_statistic}. We remove words that occur less than 5 times in the training data, and it results in 4833 words being removed. However, this reduction in term of word count is only 2.48\%. Furthermore, even though the model is evaluated in comparison to all reference sentences in the test data, there are actually 1228 words within the references that are not in our training corpus. Thus, it is impossible for the model to predict those words, and this is a limitation on scoring with references in all language models. For a better comparison with the 1000 generated captions, we also compute another reference corpus based on the first sentence of each test image. From Table \ref{tab:word_statistic}, it can be seen that even though there are at least 1187 possible words to be inferred with images in the test set, the generated descriptions are made up from only 128 and 154 words in NIC \cite{Vinyals2015} and phi-LSTM model, respectively. These numbers show that the actual number of words learned by these two models are barely 10\%, suggesting more research is necessary to improve the learning efficiency in this field. Nevertheless, it shows that introducing the phrase-based structure in sequential model still improves the diversity of caption generated.

\begin{table}[t]
\small
	\caption{Top 5 (a) least trained word found, and (b) most trained word missing, from generated caption in the Flickr8k dataset.}
		\centering
	\subtable[]{\scalebox{0.9}{
		\begin{tabular*}{0.57\linewidth}{l|c|l|c}
			\hline
			\multicolumn{2}{c|}{\textbf{NIC \cite{Vinyals2015}}} & \multicolumn{2}{c}{\textbf{phi-LSTM}} \\
			\hline
			\multicolumn{1}{c|}{Word} & Occurrence & \multicolumn{1}{c|}{Word} & Occurrence \\
			\hline
			\textit{obstacle} & 93 & \textit{overlooking} & 81 \\
			\textit{surfer} & 127 & \textit{obstacle} & 93 \\
			\textit{bird} & 148 & \textit{climber} & 96 \\
			\textit{woods} & 155 & \textit{course} & 106 \\
			\textit{snowboarder} & 166 & \textit{surfer} & 127 \\
			\hline
		\end{tabular*}
		\label{tab:word_found}
	}}
	\subtable[]{\scalebox{0.9}{
		\begin{tabular*}{0.45\linewidth}{l|c|l|c}
			\hline
			\multicolumn{2}{c|}{\textbf{NIC \cite{Vinyals2015}}} & \multicolumn{2}{c}{\textbf{phi-LSTM}} \\
			\hline
			\multicolumn{1}{c|}{Word} & Occurrence & \multicolumn{1}{c|}{Word} & Occurrence \\
			\hline
			\textit{to} & 2306 & \textit{while} & 1443 \\
			\textit{his} & 1711 & \textit{green} & 931 \\
			\textit{while} & 1443 & \textit{by} & 904 \\
			\textit{three} & 1052 & \textit{one} & 876 \\
			\textit{small} & 940 & \textit{another} & 713 \\
			\hline
		\end{tabular*}
		\label{tab:word_missing}
	}}
	\label{tab:traincount}	
\end{table}

To get further insight on how the word occurrence in the training corpus affects the word prediction when generating caption, we record the top five, most trained words that are missing from the corpus of generated captions, and the top five, least trained words that are predicted by both models when generating description, as shown in Table \ref{tab:traincount}. We consider only those words that appear in the reference sentences to ensure that these words are related to the images in the test data. It appears that the phrase-based model is able to infer more words which are less trained,  compared to the sequence model. Among the top five words that are not predicted, even though they have high occurrence in the training corpus, it can be seen that those words are either not very observable in the images, or are more probable to be described with other alternative. For example, \textit{the} is a more probable alternative of \textit{another}.

\begin{figure}[t]
	\centering
	\includegraphics[height=0.4\linewidth, width=0.9\linewidth]{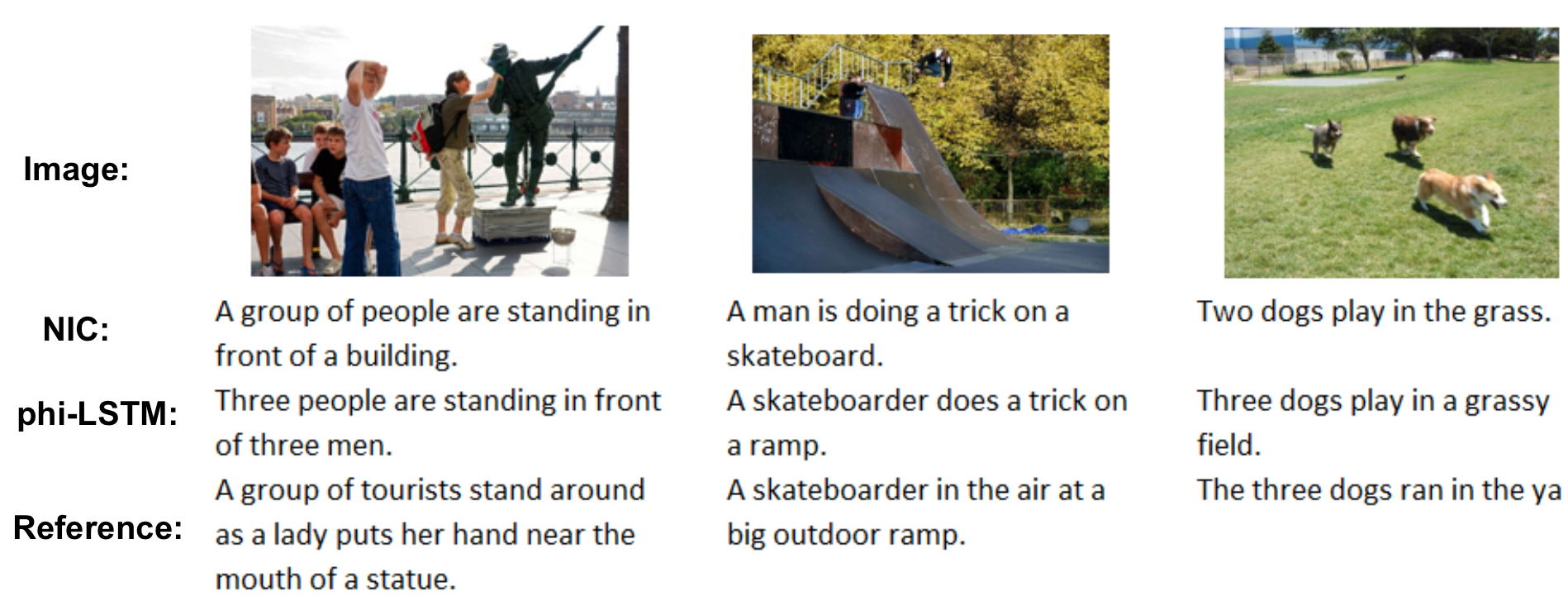}
	\caption{Examples of caption generated with phi-LSTM model, in comparison to NIC \cite{Vinyals2015}.}
	\label{fig:eg_sents}
\end{figure}

A few examples of the image description generated with our proposed model and NIC model \cite{Vinyals2015} are shown in Fig. \ref{fig:eg_sents}. It can be seen that both models are comparable qualitatively. An interesting example is shown in the first image where our model mis-recognizes the statue as a person, but is able to infer the total number of ``persons" within the image. The incorrect recognition stems from insufficient training data on the word \textit{statue} in the Flickr8k dataset, as it only occurs for 48 times, which is about 0.015\% in the training corpus.

\section{Additional Results}

\begin{figure}[H]
	\centering
	\includegraphics[height=0.65\paperheight, width=\linewidth]{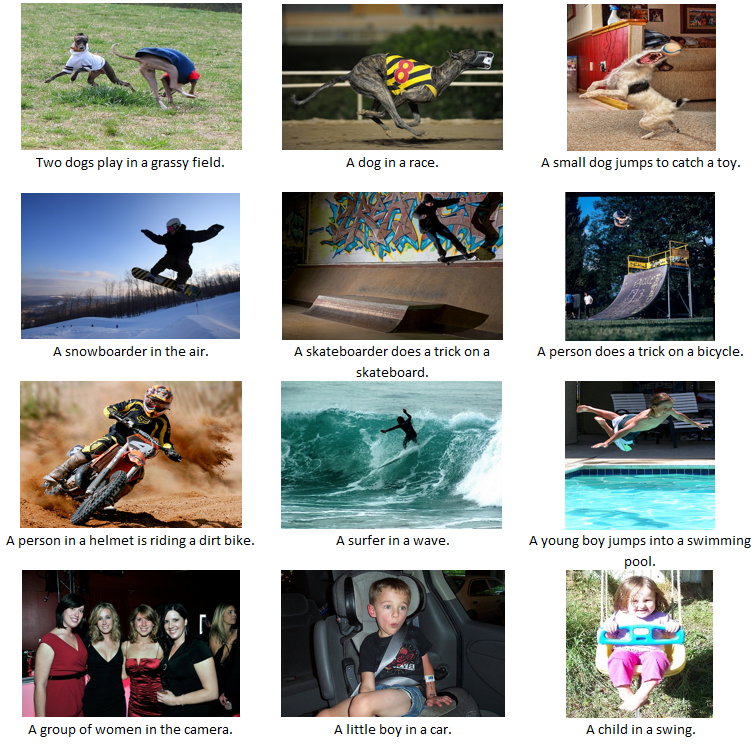}
	\caption{Flickr8k dataset: Sample image captioning results. It can be seen that our proposed model is able to distinguish different actions performed by the same subject.}
	\label{fig:f8k_results}
\end{figure}

\begin{figure}[H]
	\centering
	\includegraphics[height=0.65\paperheight, width=\linewidth]{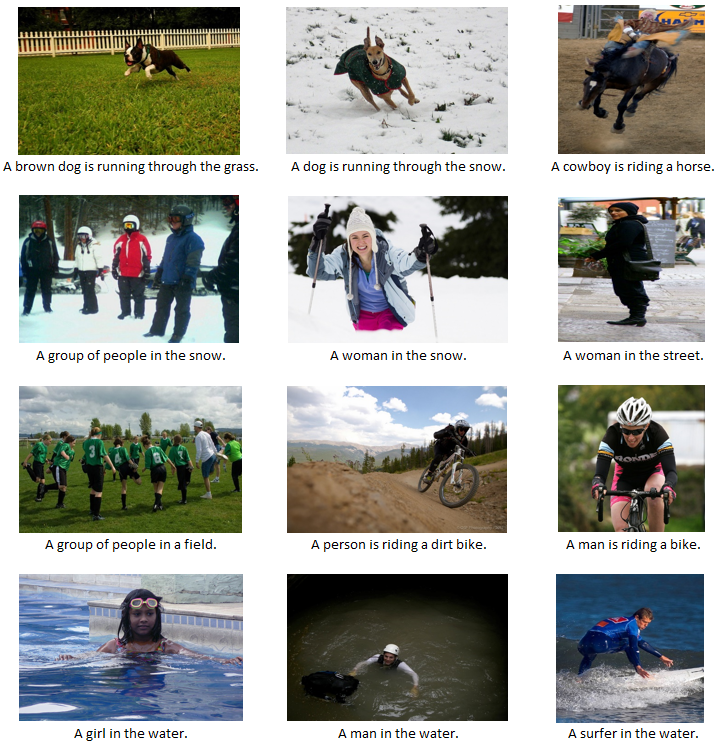}
	\caption{Flickr30k dataset: Sample image captioning results. It can be seen that our proposed model is able to distinguish different actions performed by the same subject.}
	\label{fig:f30k_results}
\end{figure}

In order to further demonstrate the capability of our proposed model - the phi-LSTM, additional results from the test set of both Flickr8k and Flickr30k datasets are shown in Fig. \ref{fig:f8k_results} and Fig. \ref{fig:f30k_results}, respectively. The results are selected such that images with very similar content are not repeatedly reported. 

Fig. \ref{fig:f8k_results} shows the outputs from the Flickr8k dataset. In the first row,  it can be seen that our proposed model is able to distinguish different actions performed by the same subject (i.e. dog), from {\it ``playing in the field"} to {\it ``racing"} to {\it ``jumping to catch a toy"}. In the second row, we demonstrate the capability of the proposed phi-LSTM model in identifying three different sports with very similar appearance in action. In particular, our model managed to detect and recognize a bicycle in the third image, even though the size of the bicycle is very small. Beside that, we also show that our proposed model is able to determine the number of subject(s) to certain extent. For example, it can identify ``\textit{two dogs}" and ``\textit{a group of women}".

Fig. \ref{fig:f30k_results} presents the outputs from the Flickr30k. Images in first row show three running actions performed by a dog and a horse in different scenes, in which the captions generated by our proposed model have correctly described them. Then, all images in the second row and the first image in the third row once again demonstrate the capability of the phi-LSTM in identifying subjects, number of subjects and scene correctly. The last two images in the third row show that our proposed model is capable of recognizing a bike, regardless the object is displayed in a partial view or a complete view. Lastly, all images in the final row display different subjects in the water, and our proposed method is able to describe each of the subjects correctly (i.e. girl, man and surfer).

Also, note that these results show that the captions generated from our proposed model are in free form, instead of fixed template like subject-verb-object or subject-action-scene. Some descriptions may describe the scene while others may not, and verb is also an optional in the description generated. The only recurring element is the subject, which is essential in the task of image description.

\begin{figure}[t]
	\centering
	\includegraphics[width=\linewidth]{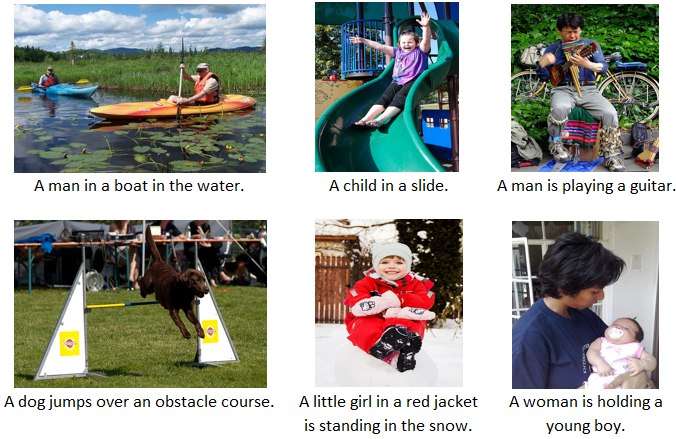}
	\caption{Examples of image captions generated with minor errors.}
	\label{fig:poor_results}
\end{figure}

Fig. \ref{fig:poor_results} shows some examples of our proposed method that have some errors in the generated captions, such as the number of subjects, actions, negligence of simultaneous action performed by subject and more specific object etc. However, it is still able to generate description that is somewhat related to the image. From our investigation, in any case, there are hardly any generated captions that infer a totally unrelated subject in the test set. 

\section{Conclusion}
In this paper, we present the phi-LSTM model, which is a neural network model trained to generate reasonable description on image. The model consists of a CNN sub-network connected to a two-hierarchical level RNN, in which the lower level encodes noun phrases relevant to the image; while the upper level learns the sequence of words describing the image, with phrases encoded in the lower level as a unit. A phrase selection objective is coupled when encoding the sentence. It is designed to aid the generation of caption from relevant phrases. This design preserves syntax of sentence better, by treating it as a sequence of phrases and words instead of a sequence of words alone. Such adaptation also splits the content to be learned by the model into two, which are stored in two sets of parameters. Thus, it can generate sentence which is more accurate and with more diverse corpus, as compared to a pure sequence model.

\vspace{3mm}
\noindent {\bf Acknowledgement}. This research is supported by the Fundamental Research Grant Scheme (FRGS) MoHE Grant from the Ministry of Higher Education Malaysia. We also would like to thank NVIDIA for the GPU donation.
 	
\newpage

\newpage

\newpage

\newpage
\bibliographystyle{splncs}
\bibliography{egbib}

\end{document}